%% file: paper.tex
\documentclass[journal]{IEEEtran}
\usepackage{cite}

\usepackage{breakurl}
\usepackage[breaklinks]{hyperref}

\usepackage{url}

\ifCLASSINFOpdf
  \usepackage[pdftex]{graphicx}
  \graphicspath{{./images/}}
\else
\fi
\usepackage{amsmath}
\usepackage{amsfonts}
\usepackage{bbm}
\usepackage{fixltx2e}

\usepackage{multirow}
\usepackage{booktabs}

\ifCLASSOPTIONcompsoc
  \usepackage[caption=false,font=normalsize,labelfont=sf,textfont=sf]{subfig}
\else
  \usepackage[caption=false,font=footnotesize]{subfig}
\fi
\begin{document}
%
\title{Cooperative Perception for 3D Object Detection in Driving Scenarios using Infrastructure Sensors}

%
%
%

\author{Eduardo~Arnold, Mehrdad~Dianati, Robert~de~Temple and Saber~Fallah
    	
\thanks{This work was supported	by Jaguar Land Rover and the U.K.-EPSRC as part of the jointly funded Towards Autonomy: Smart and Connected Control (TASCC) Programme under Grant EP/N01300X/1. 
	
	\par E. Arnold and M. Dianati are with the Warwick Manufacturing Group, University of Warwick, Coventry, UK.  (e-mail: e.arnold@warwick.ac.uk).	
	
	\par R. de Temple is with Jaguar Land Rover Ltd., Coventry, UK.
	
	\par S. Fallah is with the Connected and Autonomous Vehicles Lab (CAV-Lab), University of Surrey, Guildford, UK.

 }}

\maketitle

\input{content}




\bibliographystyle{IEEEtran}
\bibliography{paper}
%

%

%



\begin{IEEEbiographynophoto}{Eduardo Arnold}
is a PhD candidate with the Warwick Manufacturing Group (WMG) at University of Warwick, UK. He completed his B.S. degree in Electrical Engineering at Federal University of Santa Catarina (UFSC), Brazil, in 2017. He was also an exchange student at University of Surrey through the Science without Borders program in 2014. His research interests include machine learning, computer vision, connected and autonomous vehicles. He is currently working on perception for autonomous driving applications at the Intelligent Vehicles group within WMG.
\end{IEEEbiographynophoto}
\vskip -2\baselineskip plus -1fil
\begin{IEEEbiographynophoto}{Mehrdad Dianati}
is a Professor of Autonomous and Connected Vehicles at Warwick Manufacturing Group (WMG), University of Warwick, as well as, a visiting professor at 5G Innovation Centre (5GIC), University of Surrey, where he was previously a Professor. He has been involved in a number of national and international projects as the project leader and work-package leader in recent years. Prior to his academic endeavour, he have worked in the industry for more than 9 years as senior software/hardware developer and Director of R\&D. He frequently provide voluntary services to the research community in various editorial roles; for example, he has served as an associate editor for the IEEE Transactions on Vehicular Technology, IET Communications and Wiley's Journal of Wireless Communications and Mobile.
\end{IEEEbiographynophoto}
\vskip -2\baselineskip plus -1fil
\begin{IEEEbiographynophoto}{Robert de Temple}	
	is a R\&D manager and technical lead for ADAS at Jaguar Land Rover. He was first an ADAS algorithms developer, then technical lead engineer for machine learning and deep learning, now R\&D manager with a strong background in autonomous driving and deep learning. His interests and expertise lie in core AI technology and algorithms as a whole.
\end{IEEEbiographynophoto}
\vskip -2\baselineskip plus -1fil
\begin{IEEEbiographynophoto}{Saber Fallah}
	is a Senior Lecturer (Associate Professor) at the University of Surrey, a past Research Associate and Postdoctoral Research Fellow at the Waterloo Centre for Automotive Research (WatCar), University of Waterloo, Canada, and a past Research Assistant at the Concordia Centre for Advanced Vehicle Engineering (CONCAVE), Concordia University, Montreal, Canada.
	Currently, he is the director of Connected Autonomous Vehicles (CAV) lab and leading and contributing to several CAV research activities funded by the UK and European governments (e.g. EPSRC, Innovate UK, H2020) in collaboration with companies active in this domain. Dr Fallah's research has contributed significantly to the state-of-the-art research in the areas of connected autonomous vehicles and advanced driver assistance systems. 
\end{IEEEbiographynophoto}




\end{document}

%% file: content.tex
\begin{abstract}
3D object detection is a common function within the perception system of an autonomous vehicle and outputs a list of 3D bounding boxes around objects of interest.
Various 3D object detection methods have relied on fusion of different sensor modalities to overcome limitations of individual sensors.
However, occlusion, limited field-of-view and low-point density of the sensor data cannot be reliably and cost-effectively addressed by multi-modal sensing from a single point of view.
Alternatively, cooperative perception incorporates information from spatially diverse sensors distributed around the environment as a way to mitigate these limitations.
This paper proposes two schemes for cooperative 3D object detection using single modality sensors.
The early fusion scheme combines point clouds from multiple spatially diverse sensing points of view before detection.
In contrast, the late fusion scheme fuses the independently detected bounding boxes from multiple spatially diverse sensors.
We evaluate the performance of both schemes, and their hybrid combination, using a synthetic cooperative dataset created in two complex driving scenarios, a T-junction and a roundabout.
The evaluation shows that the early fusion approach outperforms late fusion by a significant margin at the cost of higher communication bandwidth.
The results demonstrate that cooperative perception can recall more than 95\% of the objects as opposed to 30\% for single-point sensing in the most challenging scenario.
To provide practical insights into the deployment of such system, we report how the number of sensors and their configuration impact the detection performance of the system.
\end{abstract}

\begin{IEEEkeywords}
	Object detection, cooperative perception, autonomous vehicles, ADAS, deep learning.
\end{IEEEkeywords}

\section{Introduction}
	\IEEEPARstart{C}{reating} an accurate representation of the driving environment is the core function of the perception system of an autonomous vehicle and is crucial for safe operation of autonomous vehicles in complex environments \cite{perceptionSurvey}.
	Failure to do so could result in tragic accidents as shown by some of the recent incidents.
	For example in two widely reported incidents by Tesla \cite{tesla-report} and Uber \cite{uber-report} autonomous vehicles, where the perception system of the subject vehicles failed to detect and classify important objects on their path in a timely manner.
	
	3D object detection is the core function of the perception system which estimates 3D bounding boxes specifying the size, 3D pose (position and orientation) and class of the objects in the environment.
	In the context of autonomous driving systems, 3D object detection is usually performed using machine learning techniques and benchmarked on established datasets such as KITTI \cite{kitti}, which provides frontal camera images, lidar point clouds and ground-truth in the form of 3D boxes annotation.
	While cameras provide rich texture information that is crucial for object classification, lidars and depth cameras produce depth information that can be used to estimate the pose of objects \cite{arnold2019survey}.
	Majority of the existing methods rely on data fusion from different sensor modalities to overcome the limitations of single sensor types and to increase detection performance \cite{mv3d,avod,fpointnet}.
	
	However, multimodal sensor data fusion from a single point of view is inherently vulnerable to a major category of sensor impairments that can indiscriminately affect various modes of sensing.
	These limitations include occlusion, restricted perception horizon due to limited field-of-view and low-point density at distant regions.
	To this end, cooperation among various agents appears to be a promising remedy for such problems. 
	While some previous studies have demonstrated limited realisations of this concept for applications such as lane selection \cite{cvLaneAssistance}, maneuver coordination \cite{maneuverCoordination} and automated intersection crossing \cite{vehicularNetworks}, this paper tackles the challenge of using the concept for cooperative perception in the form of 3D object detection, extending our previous work in \cite{coop_classification}.
	For this purpose, information from single modality, spatially diverse sensors distributed around the environment is fused as a remedy to spatial sensor impairments as alluded above.
	The benefits are many-fold: for example, observations of the environment from diverse poses increase the perception horizon, increase the density of point clouds, and hence reduce the adverse impacts of sensing noise.
	
	Cooperative perception for 3D object detection can be realised in two distinct schemes, late or early fusion, depending on whether the fusion happens after or before the object detection stage.
	In late fusion, each sensor observation is processed independently and the results in the form of detected 3D boxes from multiple sensors are fused as an end product.
	In contrast, early fusion aggregates raw sensor data and fuses it before the detection stage.
	For this reason the late fusion scheme is an example of high level fusion (object level), while early fusion is an example of low level fusion (signal level) \cite{castanedo2013review}.
	Both schemes can extend the perception horizon and field-of-view of the sensing system, however only the early fusion scheme can most effectively exploit complimentary information obtained from raw sensor observations.
	A simple illustrative example is when a vehicle is partially occluded when observed from two different sensing poses.
	In such case, each sensor observes a different occlusion pattern, resulting in unsuccessful detection from both observations. 
	In contrast, when fused, these occluded observations can provide sufficient information to successfully detect the subject vehicle.
		
	Building on our previous work \cite{coop_classification}, where we proposed the preliminary concept of cooperative perception, this paper applies the concept for cooperative 3D object detection with two distinct fusion schemes, namely \textit{late} and \textit{early fusion}, and a hybrid of the two, the so called \textit{hybrid fusion} scheme.
	We propose a system that produces a highly accurate and reliable perception of complex road segments, such as complex T-junctions and roundabouts, using a network of road-side infrastructure sensors with fixed positions.
	This perception information then can be disseminated in the form of periodic cooperative perception broadcast messages to the areas of interest in real time to assist safe autonomous driving in such areas.
	In addition to the aforementioned benefits of such system in terms of accuracy and detection performance, we believe that the proposed approach is a cost effective way of enabling safe autonomous driving systems in complex road segments.
	The main contributions of this paper can be summarised as follows:
	\begin{itemize}			
		\item Two novel cooperative 3D object detection schemes are proposed, each of them employing a distinct fusion mechanism, a bespoke deep neural network based detection and customized training procedure.
		\item A new dataset is synthesised for cooperative perception using up to eight infrastructure sensors that can be used for multi-view simultaneous 3D object detection.
		\item Comprehensive evaluations of both early and late fusion schemes, as well as their hybrid combination, are carried out in terms of detection performance and communication costs required for the operation of the system.
		\item The impacts of sensors and system configurations are analysed in order to provide insights into practical deployment of such systems.
	\end{itemize}	
	
	The rest of this paper is structured as follows.
	In Section \ref{sec:relatedworks} we review the related work in the literature and explain how our contributions differ from those.
	The proposed cooperative detection system model and the fusion schemes are explained in Section \ref{sec:method}.
	The synthesized dataset and the training process are described in Section \ref{sec:dataset} and \ref{sec:eval:train}, respectively.
	The system evaluations are presented and discussed in Section \ref{sec:experiments}.
	Finally, Section \ref{sec:conclusion} summarises the key conclusions of this paper.	
	
\section{Related works}
\label{sec:relatedworks}
	This section presents works related to 3D object detection for driving applications and data fusion schemes.
	The first subsection reviews detection models, which are designed and generally used for single sensor systems.
	The following subsection discusses the existing cooperative 3D object detection schemes in the literature, and explains how our work differs from those.
	Finally, last subsection discusses how our proposed fusion schemes fit into a taxonomy of data fusion schemes.
	
	\subsection{3D Object Detection Models}
		These models can be categorized according to the input data modality: a) colour images from monocular cameras, b) point clouds from lidar or depth cameras, or c) the combination of both.
		Monocular cameras do not provide depth cues, unless using structure from motion approaches \cite{sfm}, which require a moving camera.
		Our review in this subsection will focus on categories (b) and (c) for the reason that monocular images lack depth information and, thus, cannot be used to accurately localise objects.
		In contrast, point clouds can be used to estimate objects’ pose with significantly higher accuracy than monocular images \cite{arnold2019survey}.		
		A dedicated comprehensive review of 3D object detection techniques, covering all categories, can be found in \cite{arnold2019survey,feng2019deep}.
		
		Models in category (b) usually project the input point cloud into a Bird-Eye-View (BEV) \cite{beltran2018birdnet} or cylindrical coordinates \cite{li_vehicle_2016} to obtain a structured, fixed-size representation that can be fed to convolutional neural networks for object detection.
		After the projection and representation of the points into a fixed size input tensor, convolutional layers are applied to generate the final 3D bounding boxes on a single forward pass \cite{Simony_complexYOLO}.
		Both projection techniques can result in loss of information due to space quantization and the representation choice.
		In contrast to projection techniques, Voxelnet \cite{voxelnet} learns a representation from the raw 3D points to obtain a structured input; PointRCNN \cite{pointrcnn} and STD \cite{std2019} are two stage detectors that use PointNet \cite{qipointnet} as a backbone to obtain point-wise 3D proposals, then refine the proposals using a specialised network.
		In the case of \cite{pointrcnn} the refinement network takes into account semantic features and local spatial features, while as in \cite{std2019} the refinement is guided by local spatial features and an IOU estimation branch.
		
		Category (c) models such as	Multi-View 3D (MV3D) \cite{mv3d} use the BEV projection of the point cloud to produce object proposals and later fuse the lidar front-view projection as well as the colour image features.
		Another example is the Aggregate View Object Detection (AVOD) model \cite{avod} that uses both the colour image and the BEV projection to generate object proposals and then fine-tunes them to obtain the final detection boxes.
		Different from \cite{mv3d,avod}, authors in \cite{fpointnet} use another fusion strategy: they firstly detect the objects on the image plane (2D bounding boxes), then extend each 2D detection into a 3D frustum and select the lidar points within the frustum as input to a PointNet model \cite{qipointnet} which segments background points and regresses a 3D bounding box to fit the segmented points.

		Building on the previous works on 3D object detection from a single point of view summarised in this subsection, this paper focuses on cooperative object detection.
		We particularly use the Voxelnet object detection model \cite{voxelnet} due to its generalisation capacity and detection performance.
		Since this model operates exclusively on point clouds, it enables us to reduce the required bandwidth for data transmission from sensor nodes to the fusion system and avoid potential privacy issues that arise with colour images.

	\subsection{Cooperative 3D Object Detection Schemes}
		Chen \textit{et al.} propose to fuse raw point clouds and introduce a neural network architecture for object detection in sparse point clouds.
		Their study \cite{chen_cooper_2019} consider communication costs, robustness to localisation error and show that cooperative perception can enhance the performance of object detection in terms of the number of detected objects.
		An extended study in \cite{fcooper} propose feature-level fusion schemes and analyse the trade-off between processing time, bandwidth usage and detection performance.
		Both works \cite{chen_cooper_2019,fcooper} used the KITTI dataset \cite{kitti}, merging two sequential frames to simulate a cooperative dataset, and their own dataset obtained with two vehicles on a parking lot.
		Hurl \textit{et al.} study the problem of trust in cooperative 3D object detection and propose TruPercept \cite{trupercept} to prevent malicious attacks against cooperative perception systems. 
		They evaluate their scheme using a cooperative synthetic dataset generated by a game engine in general urban scenarios.
		
		Our study differs from the aforementioned works in three main aspects.
		Firstly, while \cite{chen_cooper_2019} and \cite{trupercept} share on-board information peer-to-peer (V2V) and fuse data locally, we propose a central system that fuses data from multiple infrastructure sensors which allows to amortize both sensor and processing costs through shared resources.
		Secondly, unlike \cite{chen_cooper_2019} and \cite{fcooper} where the authors evaluated their system on a few scenes from the KITTI dataset and a parking lot scenario, we tackle two complex urban scenarios, a T-junction and a roundabout, where occlusion is most severe.
		Thirdly, our study addresses evaluations of practical aspects of sensor configurations such as the number of sensors, their pose and field-of-view overlapping, which have been overlooked by the aforementioned works and provide important practical insights into the deployment of such systems.
		
	\subsection{Data Fusion Schemes}
		Castanedo \cite{castanedo2013review} presents a taxonomy for different data fusion schemes and reviews the most common techniques within three categories: data association, state estimation and decision function.
		Regarding the relationship between data sources the presented taxonomy allows to classify both our early and late fusion schemes as complementary, when the sensors provide exclusive field-of-views, and as redundant, when using sensors with overlapping field-of-views.
		
		Ghamisi \textit{et al.} \cite{ghamisi2019multisource} reviews three categories of point cloud fusion for remote sensing: point cloud level, where more points or features are added to the initial point cloud; voxel level where point clouds are fused in a voxel representation; and feature level, where features are fused on the object level.
		Our early fusion scheme is within the first category, where an existing point cloud is extended with points from spatial diverse sensors of the same modality.
		However our late fusion scheme does not fit any of the categories since it fuses the detected bounding boxes themselves, rather than points, voxels or point cloud features.						
		
\section{System Model and Fusion Schemes}
\label{sec:method}
	We firstly describe our system model for all fusion schemes in Section \ref{sec:method:sysmodel}.
	Section \ref{sec:method:preprocessing} presents the data preprocessing stage, while the proposed early, late and hybrid fusion schemes for cooperative 3D object detection are described in Sections \ref{sec:method:efusion}, \ref{sec:method:lfusion} and \ref{sec:method:hfusion}, respectively.
	Finally, Section \ref{sec:method:model} introduces the 3D object detection model used in our system.
	
	\begin{figure}
		\centering
		\includegraphics[width=\linewidth]{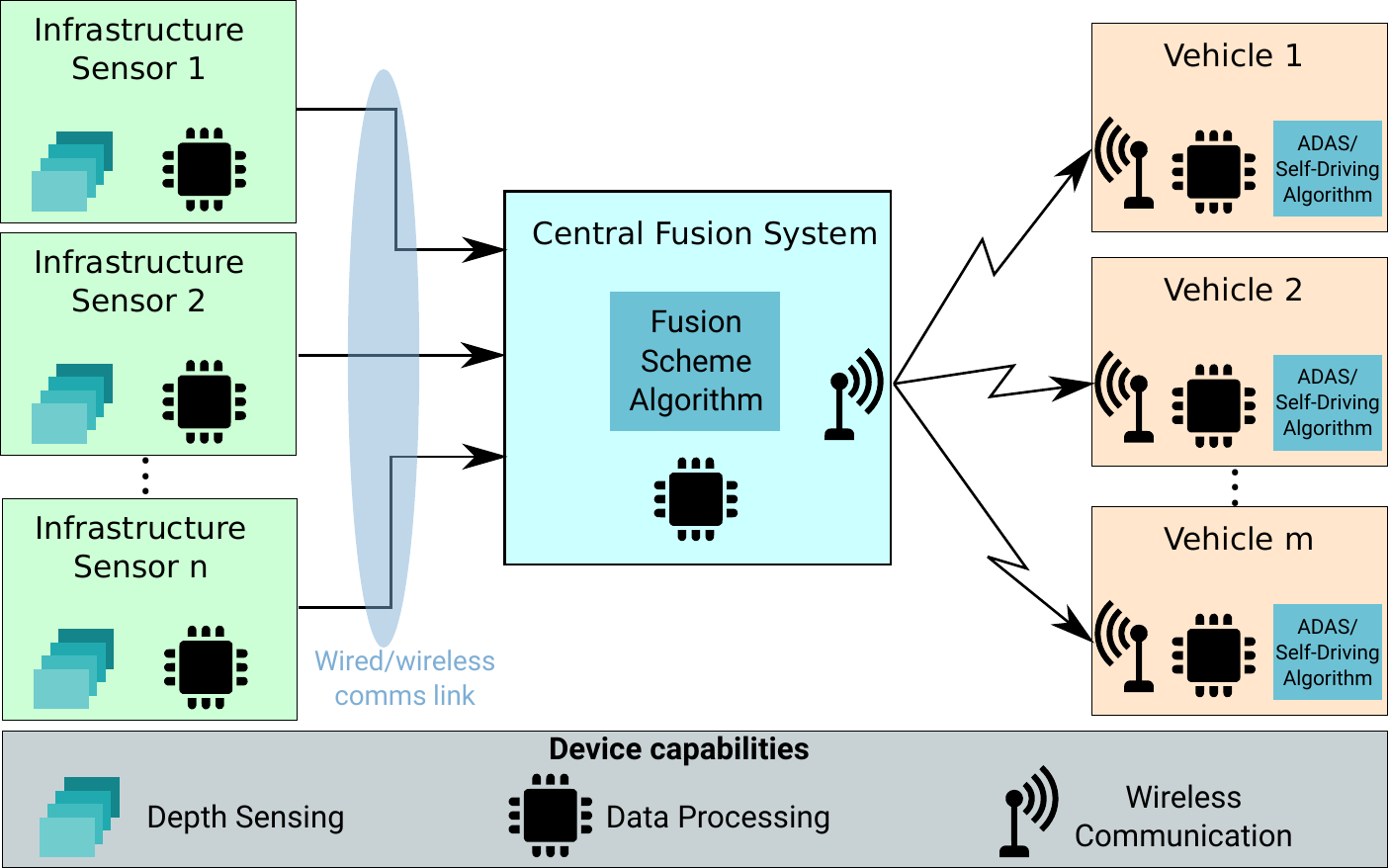}
		\caption{Cooperative 3D Object Detection System Model. The data provided by sensors is fused at the central fusion system resulting in a list of objects which is then shared with all nearby vehicles.}
		\label{fig:logical}
	\end{figure}
	
	\begin{figure*}
		\centering
		\includegraphics[width=\linewidth]{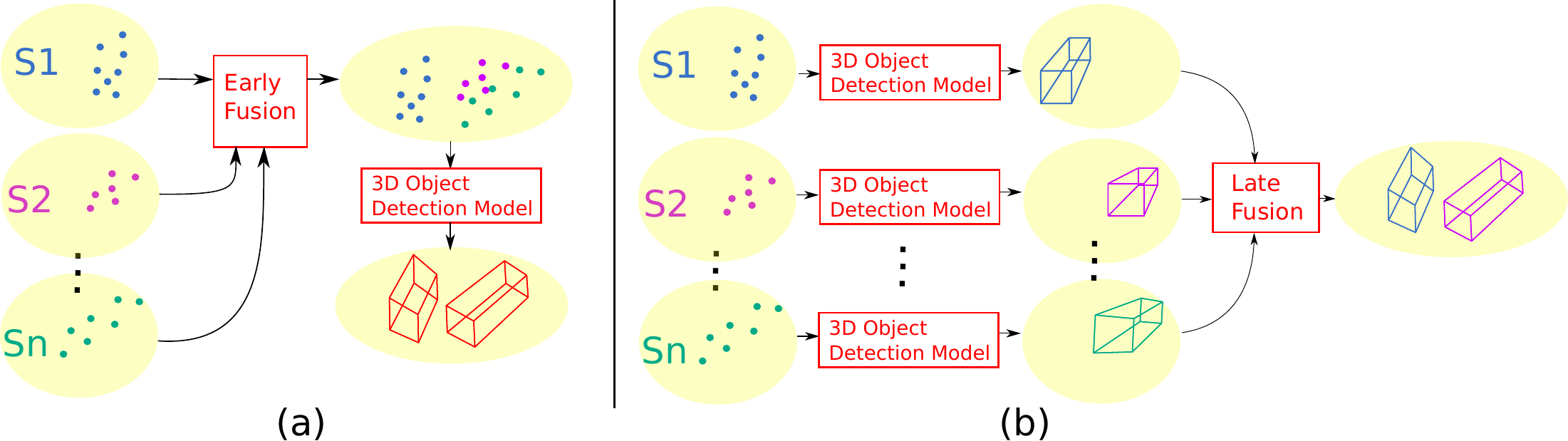}
		\caption{Logical illustration of Early (a) and Late (b) schemes for Cooperative 3D Object Detection.} 
		\label{fig:fusion}
	\end{figure*}
	
	\subsection{System Model}
	\label{sec:method:sysmodel}	
		As shown in Figure \ref{fig:logical}, our proposed system model considers $n$ infrastructure sensors, each capable of depth sensing, \textit{e.g.} lidar or depth camera, and equipped with a local processor.
		These infrastructure sensors are linked to a central fusion subsystem through wired or wireless data links.
		The sensors are assumed to be accurately calibrated, \textit{i.e.} their absolute pose, including position and orientation, is known to the central system.
		The central fusion subsystem is equipped with a fairly powerful processor to fuse data from sensors and a wireless broadcast system to periodically disseminate cooperative perception messages to the vehicles in the proximity.
		To benefit from the cooperative perception messages, each vehicle will need to be equipped with a wireless reception system.
		Autonomous vehicles also are assumed to have their own onboard processing system to handle the local processing of either Advanced Driver Assistance Systems (ADAS) or autonomous driving functions, including perception and control (\textit{e.g.} path/trajectory planning).
		It shall be noted that the central fusion subsystem is not responsible for the control of the vehicles.
		Each autonomous vehicle will therefore use on and off-board (broadcast by the central fusion subsystem) information to make their own control decisions.
		Hence, in our system model, the role of the central fusion system is only to assist the vehicles in making safer control decisions.
		We would like to note that the network delay and communication losses are not considered in this paper and will be taken into account in future studies.			
	
	\subsection{Data Preprocessing}
	\label{sec:method:preprocessing}	
		The detection model considered in this paper requires point cloud data, such as that provided by lidars or depth cameras. 
		While lidars can produce point clouds as a standard output, the depth images produced by depth cameras can be processed (details in Section \ref{sec:dataset}) to generate point clouds.
		Each sensor in a physical configuration provides points relative to its own coordinate system, thus they need to be transformed to a global coordinate system before being processed.
		This transformation consists of a rotation and a translation operation that maps points from the sensor coordinate system to a global coordinate system and is specified by the inverse of the extrinsic matrix of each sensor.
		Given the coordinates $(x,y,z)$ of a point in the coordinate system of sensor $i$, we can obtain the global reference point $(x_g,y_g,z_g)$ using:
		\begin{equation}
			\label{eq:tr-cs}
			\begin{bmatrix}
			x_g \\
			y_g \\
			z_g \\
			1
			\end{bmatrix} = M_i^{-1}
			\begin{bmatrix}
			x \\
			y \\
			z \\ 
			1			
			\end{bmatrix} = 
			[R_i | t_i]		
			\begin{bmatrix}
			x \\
			y \\
			z \\ 
			1			
			\end{bmatrix},
		\end{equation}
		where $M_i$ is the extrinsic matrix of sensor $i$, which can be decomposed into a rotation matrix $R_i$ and translation vector $t_i$.
		
		The extrinsic matrix $M$ of a sensor, and hence $R$ and $t$, must be obtained through a calibration process. 
		This can be challenging in practice for the reason that $M$ depends on the position and orientation of sensors; hence, the result can only be as accurate as the measurements of these variables.
		Realistically, if these sensors are mounted on mobile nodes (\textit{e.g.} onboard of a vehicle) any error in the localisation of the mobile node will result in alignment errors in the fused point cloud which could result in false positives and missed detections.
		In our system model the sensors are fixed at the road side; therefore, the calibration process can be carried out very accurately in practice \cite{yue2019data,cameracalibration}.
		
		Once the point cloud from each sensor is transformed into the global coordinate system, all the points outside the specified detection area (defined in Section \ref{sec:dataset}) and above 4m height are removed since such points do not carry relevant information.
	
	\subsection{Early Fusion Scheme}
	\label{sec:method:efusion}
		This scheme, as illustrated in Figure \ref{fig:fusion}a, is based on the fusion of point clouds generated by $n$ sensors, as depicted in Figure \ref{fig:logical}.
		This allows aggregation of complementary information from distinct parts of the objects in the detection area through spatially diverse observations, which increases the likelihood of a successful detection, particularly for objects that are occluded or have low visibility.		
		The processing pipeline for this scheme incorporates the preprocessing stage carried out onboard of each sensor, which results in $n$ point clouds in the global coordinate system.
		Each respective point cloud is transmitted to the central fusion system where they are concatenated into a single point cloud and then fed to the 3D object detection model.
		The results of the object detection model in the central fusion system consists of a list of objects, \textit{i.e.} 3D bounding boxes, which is then disseminated to the vehicles in the vicinity, as depicted in Figure \ref{fig:logical}.

	\subsection{Late Fusion Scheme}
	\label{sec:method:lfusion}
		This scheme, as illustrated in Figure \ref{fig:fusion}b, fuses the output of the 3D object detection model (a list of 3D bounding boxes) obtained locally at each sensor node.
		Thus, if an object is not detected in at least one of the observed point clouds, \textit{e.g.} due to occlusion or low point density, it cannot be detected by the overall system.		
		First, each point cloud is preprocessed and fed into the detection model onboard each sensor, which generates a list of objects represented by their 3D bounding boxes.
		The list of detected objects from the $n$ sensors are then transmitted to the central fusion system, where they are fused into a single list.
		Considering that some objects may be in the field-of-view of multiple sensors, the aggregated list may have multiple detections for a single object.
		In order to mitigate this effect, we use a post-processing algorithm known as Non-Maximum Suppression (NMS) \cite{felzenszwalb2009object}.
		This algorithm identifies the overlap of the detected boxes, measured by the Intersection Over Union (IOU) metric (described in Section \ref{sec:experiments}).
		If the overlap between any two detected boxes exceeds a specified threshold, the box with lowest confidence score is removed.
		The confidence score of a detected box indicates the confidence of the presence of an object within the box, and is obtained by the 3D object detection model (detailed in Section \ref{sec:method:model}).
		Figure \ref{fig:fusion}b illustrates an example case where S2 and Sn observations resulted in two detections of a single object, thus, during the fusion stage the box detected by Sn is omitted.
		A number of detected boxes that overlap could be potentially combined to create a new detection box with higher confidence, however this would require a new model specifying the box fusion process.
		The conducted experiments showed that NMS was successful in eliminating the overlapping detected boxes and thus we opted to use this algorithm for simplicity.
		Once the fusion and post-processing are completed, the resulting object list is broadcast to all vehicles in the vicinity, similar to the previous scheme.
		
	\subsection{Hybrid Fusion Scheme}
	\label{sec:method:hfusion}
		The early fusion scheme can increase the likelihood of detecting objects compared to late fusion due to the aggregated information prior to the detection stage but requires raw sensor data sharing, which increases the communication cost.		
		As an intermediate solution, the hybrid fusion scheme uses both of the previous schemes to increase the likelihood of a detection without a drastic increase in the communication cost.
		The key concept is to share high level information (late fusion) where the sensor has high visibility and share low level information (early fusion) where the visibility is poor.
		Objects close to a sensor will have a high density of points and thus are more likely to be detected using a single sensor's observation.
		Thus points in the close vicinity of a sensor need not be transmitted to the central fusion system, which allows to reduce the communication bandwidth.
		First, the late fusion scheme is employed in each sensor node and the detected boxes are shared to the central fusion system.
		Next, each sensor node selects all points from its point cloud whose projection in the horizontal plane are outside a circle of radius $R$ and share them with the central fusion system.
		The radius $R$ modulates the trade-off between early and late fusion -- as $R$ decreases more raw data is shared with the central fusion system.
		The central fusion system then uses early fusion on the received point clouds and fuses the detected bounding boxes with the late fusion results from each sensor node.
		The bounding box fusion follows the same NMS procedure defined in Section \ref{sec:method:lfusion}.				
		
	\subsection{3D Object Detection Model}
	\label{sec:method:model}
		The object detection model adopted in this paper for all fusion schemes is based on Voxelnet \cite{voxelnet}.
        This model consists of three main functional blocks: a feature learning network, multiple convolutional middle layers and a Region Proposal Network (RPN).
		Each block is described below.
		
        \begin{figure*}
	       \centering
	       \includegraphics[width=0.8\linewidth]{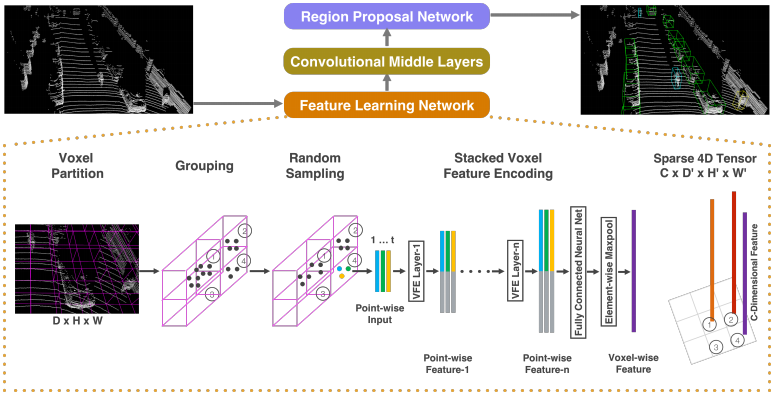}
	       \caption{Voxelnet 3D object detection model architecture. Image from \cite{voxelnet}.}
	       \label{fig:voxelnet}
       \end{figure*}
		
		The feature learning network converts the 3D point cloud data into a fixed sized representation that can be processed by the convolutional layers.
		Originally, the Voxelnet architecture used the laser reflection intensity channel as well as the 3D spatial coordinates $(x,y,z)$.
		In our implementation, we use the spatial coordinates alone, which enables the model to generalise to point clouds obtained from depth cameras.
		The input point cloud is grouped into voxels of equal size ($v_x$,$v_y$,$v_z$), representing the width, length and height, respectively.
		For each voxel, a set of $t$ points within its boundaries is selected to create a voxel-wise feature vector.
		If $t$ is greater than $T$ (threshold on the maximum number of points per voxel), a random sample of $T$ points is selected, which reduces the computational load and the imbalance of the number
		of points between different voxels.
		These points' coordinates are fed into a chain of Voxel Feature Encoding (VFE) layers.
		Each VFE layer in the chain consists of fully connected layers, local aggregations and max-pooling operations that allow concentration of information from all voxel points into a single voxel-wi
		se feature vector.
		The output of this network is a 4D tensor, indexed by the voxel feature dimension, height, length and width.
		
		The convolutional middle layers in the processing pipeline apply three stages of 3D convolutions to the 4D voxel tensor obtained previously.
		These stages incorporate information from neighbouring voxels, adding spatial context to the feature map.
		
		The resulting tensor from the convolutional middle layers is then fed into the Region Proposal Network.
		This network is composed of three stages of convolutional layers, followed by three stages of transposed convolutional layers which create a high resolution feature map.
		This feature map is then used to generate two output branches: a confidence score indicating the probability of presence of an object and a regression map indicating the relative size, position,
		and orientation of a bounding box with respect to an anchor.
		Each element of the output branch is mapped to an anchor in a uniformly arranged grid, whose density is controlled by the anchor stride hyper-parameter.
		Anchors are used since the regression of detection boxes relative to an anchor gives more accurate results than regression without any prior information \cite{ren2015faster}.

\section{Dataset}
\label{sec:dataset}	
	To the best of our knowledge there is no publicly available dataset that can be readily used for cooperative 3D object detection in the literature.
	We would like to note that authors in \cite{chen_cooper_2019} and \cite{fcooper} simulate an environment where two vehicles share their sensors' information by using point clouds from the KITTI dataset \cite{kitti} generated by the same vehicle at two time instants.
	However, their approach is limited to a small number of scenes where all objects remain static, except for the ego vehicle, which also restrict the number and pose of sensors that can be used.
	Hence, it falls short of providing a comprehensive dataset to investigate dynamic and complex driving scenarios where multiple objects are moving and/or are occluded.
	To this end, we generate a novel cooperative dataset for driving scenarios using multiple infrastructure sensors as described in the following.
	
	Our dataset was created using the CARLA simulation tool \cite{carla}, which allowed simulation of complex driving scenarios as well as obtaining accurate ground-truth data for training and evaluation.
	This dataset is used in our paper to establish the underlying concepts of our cooperative 3D object detection schemes and gauge the potential benefits.
	In the next phase of our research, we plan to use realistic datasets generated from our outdoor track which is currently under development as part of the Midlands Future Mobility testbed \cite{MFMwebsite}.
	
\begin{figure*}[htp]
		\centering
		\subfloat[\label{fig:config-t}]{\includegraphics[width=0.48\textwidth]{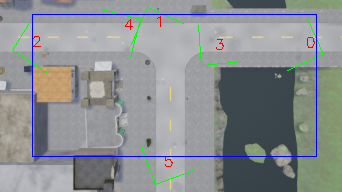}}\hfill
		\subfloat[\label{fig:pcl-t}]{\includegraphics[width=0.48\textwidth]{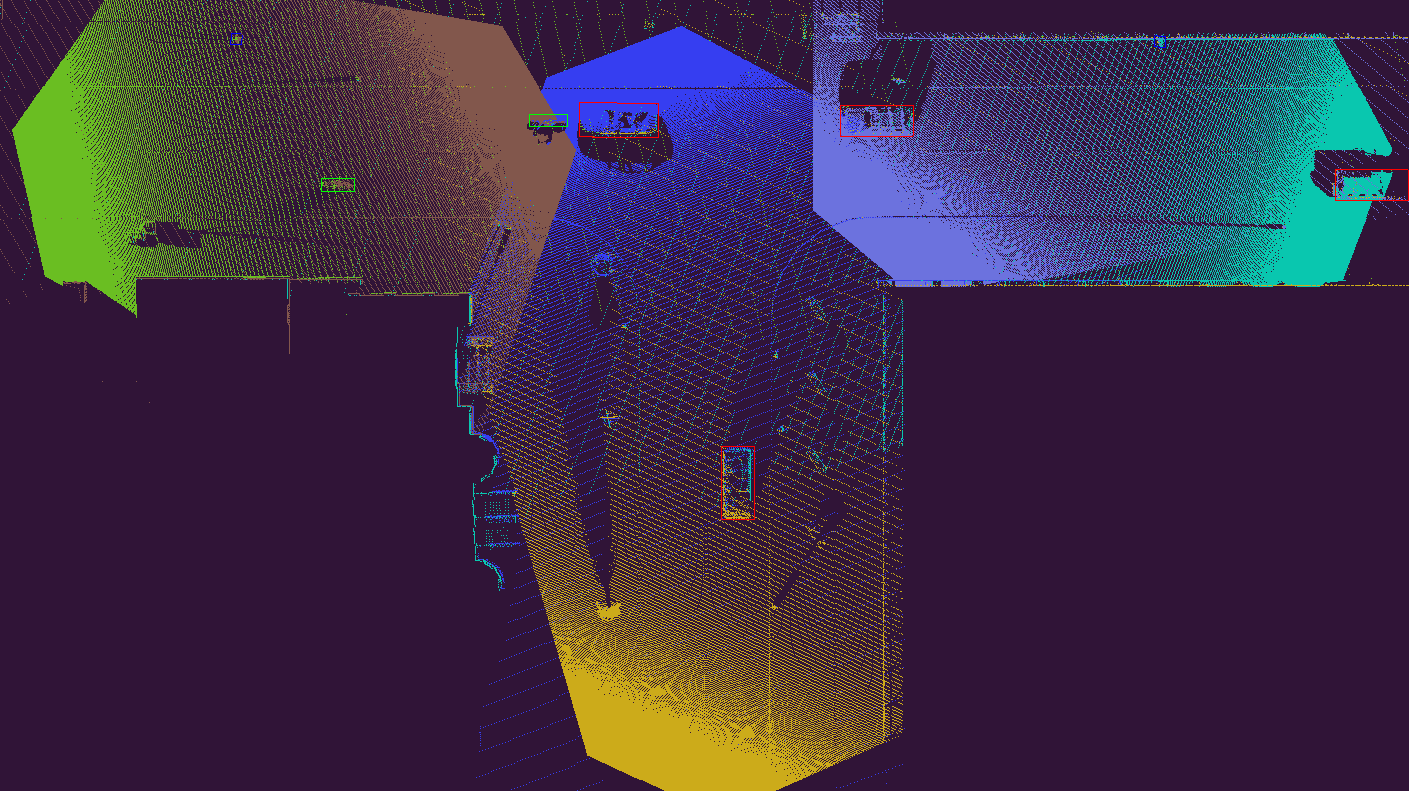}}

		\subfloat[\label{fig:config-r}]{\includegraphics[width=0.48\textwidth]{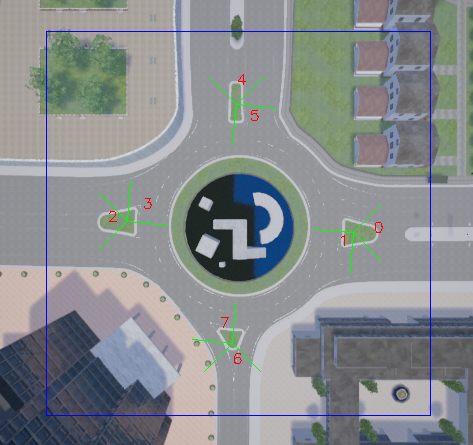}}\hfill
		\subfloat[\label{fig:pcl-r}]{\includegraphics[width=0.48\textwidth]{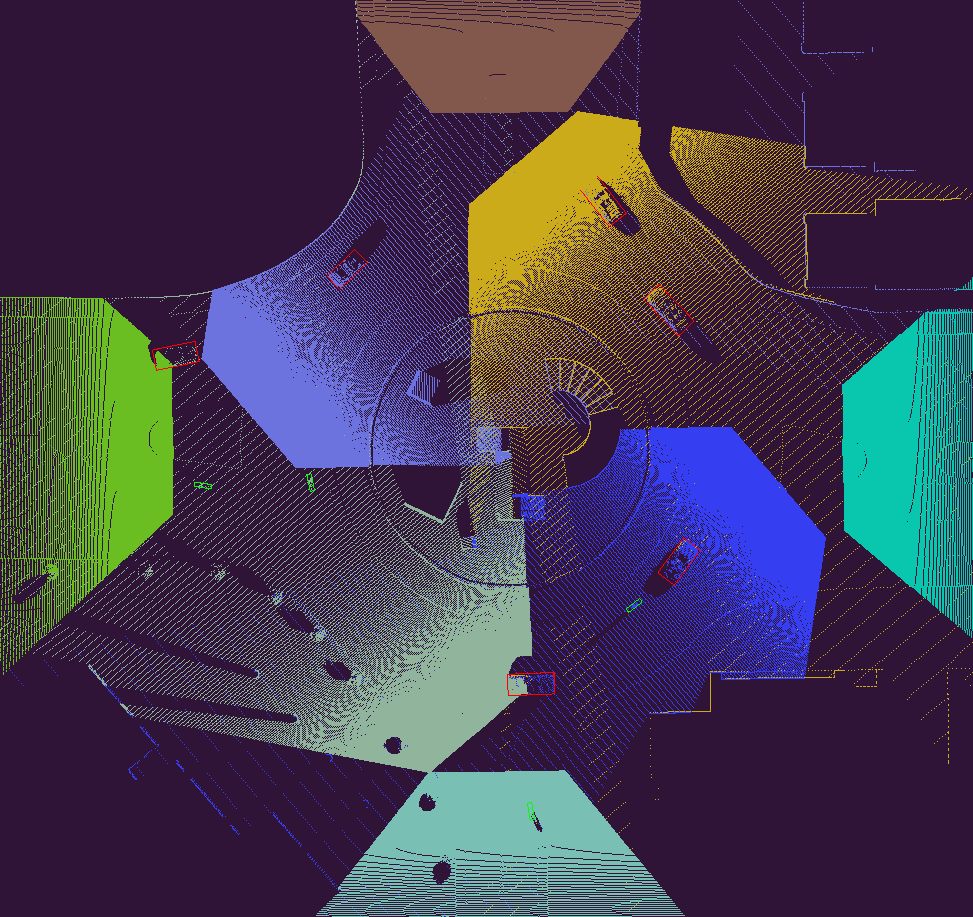}}		
		
		\caption{Bird Eye View of the T-junction and Roundabout scenarios. \protect\subref{fig:config-t} and \protect\subref{fig:config-r} show the sensors configuration, where each sensor is indicated by its ID number and green lines representing the field-of-view; the blue rectangles indicate the detection areas of each scenario. \protect\subref{fig:pcl-t} and \protect\subref{fig:pcl-r} show the fused point clouds, where each colour represents a sensor and the 3D bounding boxes represent the labelled data, with colour indication of the class (red for cars, green for cyclists and blue for pedestrians). Note that this sensor configuration fully covers the detection areas and provide overlapping field-of-views (indicated by areas with multiple colours).}
		\label{fig:dataset}
	\end{figure*}

	Our dataset is generated in a T-junction and a roundabout scenario using fixed road-side cameras, which provide RGB and depth images with resolution of 400 x 300 pixels and horizontal field-of-view of 90 degrees.
	The resolution and field-of-view are conservative estimates of new generation solid state lidars, whose specifications are not yet available \cite{velarray}.
	The T-junction scenario uses six infrastructure cameras mounted on 5.2m high posts.
	Three of these cameras point towards the incoming roads and the remaining three to the opposite direction of the junction.
	The roundabout scenario uses eight cameras at 8m mounting posts placed at the intersections, four of them facing the oncoming lanes to the roundabout and the other four facing outwards the roundabout.
	The sensors' height was chosen according to the typical light poles height already available in the simulation scenarios and to conform to local UK standards \cite{durhamLighting}.
	Both sensor configurations were experimentally positioned to fully cover the roundabout and junction, as illustrated in Figure \ref{fig:dataset}.	
	
	The proposed dataset consists of four independent collections: two for the T-junction, containing 4000 training and 1000 test frames respectively, and two for the roundabout, with an equal number of training and test frames.
	A frame is defined as the set of depth and RGB images from all cameras corresponding to a single instant in time.
	Each frame also contains an object list describing the ground-truth position, orientation, size and class of all objects in the scene.
	
	The objects represented in the dataset can be vehicles, cyclists/motorcyclists or pedestrians.
	Note that we do not distinguish between cyclists and motorcyclists in this paper.
	During the generation of the dataset, the maximum number of objects at any given time was set to 30, which was observed as a threshold above which severe traffic congestion happens.
	The probabilities of spawning cars, cyclists and pedestrians is equal to 0.6, 0.2 and 0.2, respectively, which guarantees a higher number of cars but still allows a representative sample of cyclists and pedestrians.
	During the simulation, each object has a life span of four frames, which forces new objects to be spawned periodically and increase the diversity of objects and poses.
	The motion of the objects in the simulation is governed by traffic rules and internal collision avoidance mechanisms of the simulator.
	All object models available in CARLA are used during the simulation -- twenty for cars (sport, vans and SUVs), six for cyclists and fourteen for pedestrians.
		
	We define the detection areas as a rectangle of 80 x 40m for the T-junction scenario and a square of 96m centred at the roundabout, illustrated by the blue rectangles in Figure \ref{fig:config-t}, \ref{fig:config-r}.
	The areas of interest for object detection are chosen to cover all the junction/roundabout area and some extent of the roads leading to it in order to increase the perception horizon of the system, while taking in account the constraints in the processing system memory.
	The T-junction and roundabout scenarios detection areas cover 3200~m$^2$ and 9216~m$^2$, respectively. 
	
	In our approach, we use only the depth images, also known as depth maps.
	These images represent the distance from the camera to the surface of objects in the camera field-of-view. 
	More accurately, each pixel in a depth image specifies the distance of the projection (into the camera's Z axis) of the vector from the camera to a surface point.
	Each depth image is used to reconstruct a point cloud, where each pixel is transformed into a 3D point in the camera coordinate system using the pinhole camera model \cite{hartley_zisserman_2004}, described by:
	\begin{equation}
		\label{eq:pcl}
		\begin{bmatrix}
			x \\
			y \\
			z
		\end{bmatrix} =
		\begin{bmatrix}
			(u-C_u)\frac{d}{f} \\
			(v-C_v)\frac{d}{f} \\
			d			
		\end{bmatrix},
	\end{equation}
	where $(x,y,z)$ are the coordinates of the 3D point corresponding to pixel coordinates $(u,v)$ in the depth image, $C_u,C_v,f$ are the camera focal centre and length (given by the intrinsic camera matrix), and $d$ is the respective depth value of pixel with coordinates $(u,v)$.	
	The point cloud produced by combining the 3D points from all cameras should have a similar size as one produced by a standard lidar, around 200 thousand points, for processing time constraints.
	To this end, the depth image resolution is downsampled in half to 200 x 150 pixels, which yields 30000 3D points per camera, and approximately 200 thousand points when combining points from six or eight cameras.
		
	We introduce a surface agnostic Additive White Gaussian Noise (AWGN) model with mean $\mu=0$m and standard deviation $\sigma=0.015$m to the depth image, following the specification and mathematical model of a lidar sensor in \cite{glennie2010static}.
	It must be noted that, in contrast to lidar sensors, the depth estimation error of stereo-matching-based cameras increases exponentially with the distance between the camera and the object \cite{ortiz2018depth}.
	
\section{Training process}
\label{sec:eval:train}
	The model described in Section \ref{sec:method:model} is trained using the procedure presented below.
	We train one instance of the 3D object detection model for each scenario using fused point clouds from multiple sensors.
	The models are trained with Stochastic Gradient Descent (SGD) optimisation for 30 epochs with learning rate of $10^{-3}$ and momentum of $0.9$, as proposed in \cite{voxelnet}.
	The loss function is adopted from \cite{voxelnet}, and penalises the regression of position, size and yaw angle relative to a fixed anchor.
	We opt for the hyper-parameters suggested in \cite{voxelnet}: a single anchor of size $(3.9,1.6,1.56)$m with two orientations (0 and 90 degrees) and maximum number of points per voxel $T=35$.
	
	The voxel size $(v_x, v_y, v_z)$ and the anchor stride along the X and Y dimensions for the T-junction model is set to $(0.2,0.2,0.4)$m and $0.4$m, respectively, identical to those in \cite{voxelnet}.
	Using these same hyper-parameters in the roundabout model is unfeasible since the roundabout scenario has approximately thrice the area of the T-junction scenario, which would result in feature maps that do not fit in the GPU memory.
	Hence, we reduce the spatial resolution of the X and Y axis in half by adopting a voxel size of $(0.4,0.4,0.4)$m and anchor stride $0.8$m for the roundabout model.
	
	The object detection models are trained to detect vehicles only.
	The samples of pedestrians and cyclists are present in the dataset to avoid over-fitting as they force the model to learn distinct features for vehicles.
	
	During the training stage, each ground-truth bounding box is rotated by a random angle with a uniform distribution in the range of $[-18,18]$ degrees, similar to previous studies in \cite{voxelnet,yan_second}, to increase the generalisation of angle estimation.	
	We also consider rotating the whole point cloud to avoid model over-fitting to the buildings and fixed objects surrounding the junction, however this operation did not result in a significant performance gain.	

\section{Performance Evaluation}
\label{sec:experiments}
	We evaluate the performance of the proposed cooperative perception system for 3D object detection through a series of experiments in two scenarios, a T-junction and a roundabout.
	The performance evaluation is carried out on an independent test dataset for each scenario using the metrics described in Section \ref{sec:eval:metrics}.
	First, we compare the fusion schemes in terms of their detection performance, computation time and communication costs for data sharing in Section \ref{sec:eval:earlyVlate}.
	Secondly, we evaluate the impact of the number of infrastructure sensors and their pose on the detection performance in Section \ref{sec:eval:sconfig}.
	Then, the benefits of fusing information from multiple sensors with overlapping field-of-view in early fusion scheme are evaluated in Section \ref{sec:eval:overlap}.
	Additionally, it is evaluated how the number of infrastructure sensors relates to the quality of the information acquired from the objects (in terms of the density of points in the point cloud data), and, in turn, how this number relates with the accuracy of the detected boxes in Section \ref{sec:eval:cov}.
	Finally, we compare our system performance to existing benchmarks in Section \ref{sec:eval:comp}.
	
	\subsection{Evaluation Metrics}
	\label{sec:eval:metrics}
		Four evaluation metrics related to object detection are used in this paper, namely, Intersection Over Union (IOU), precision, recall and average precision, which is derived from the previous two.
		Additionally, the communication cost metric is defined as the average data volume exchanged between a sensor and the central fusion system in \textit{kilobits (kbit) per frame}, where a frame is defined as a single operation of the whole processing chain in this paper.
		
		The IOU measures the spatial similarity of a pair of bounding boxes, one normally chosen from the set of estimated bounding boxes and the other from the ground-truth set, given by
		\begin{equation}
		\label{eq:iou}
			\text{IOU}(B_{gt}, B_e) = \frac{\text{volume}(B_{gt} \cap B_e)}{\text{volume}(B_{gt} \cup B_e)},
		\end{equation}
		where $B_{gt}$ and $B_e$ represent the ground-truth and the estimated bounding boxes, respectively.
		The set of estimated bounding boxes includes all positive boxes, \textit{i.e.}, those identified by the 3D object detection model in Section \ref{sec:method:model} with confidence scores greater than a threshold, denoted by $\tau$ in this paper.
		The IOU simultaneously takes into account the location, size, and orientation (yaw angle) of both bounding boxes.
		Its value ranges from 0 (when the bounding boxes do not intersect) to 1 (when the location, size, and orientation of both bounding boxes are equal).
		Normally, when the IOU metric for a pair $(B_{gt}, B_e)$ is above a certain threshold, denoted by $\kappa$, $B_e$ can be regarded as the matching estimation of $B_{gt}$.
		The IOU threshold $\kappa$ is typically set to 0.5 or 0.7 \cite{kitti}, here we opt for 0.7 unless stated otherwise.

		The precision metric is defined as the ratio of the number of matched estimated boxes, according to the above definition, to the total number of bounding boxes in the estimated set.
		Similarly, the recall metric is defined as the ratio of the number of matched estimated boxes to the total number of bounding boxes in the ground-truth set.
		It shall be noted that the precision and recall metrics are functions of $\kappa$ and $\tau$.
		And, there is an inherent trade off between the precision and recall metrics, described in the literature by the Precision-Recall (PR) curve \cite{voc2010pascal}.

		\newcommand{\ap}{$\textsc{AP}_{\textsc{3D}}$}
		\newcommand{\aproi}{$\textsc{AP}^{\textsc{ROI}}_{\textsc{3D}}$}
		
		The Average Precision (AP), denoted as \ap{}, is a single scalar value, computer by taking the average of the precision for $M$ recall levels \cite{voc2010pascal,vocRetrospective}:
	 	\begin{equation}
	 	\label{eq:ap}
			\text{AP} = \sum_{n=0}^{M-1} (r_{n+1}-r_n)p_\text{interp}(r_{n+1}),
		\end{equation}
		where 
		\begin{equation}
			p_\text{interp}(r) = \max_{\tilde{r}:\tilde{r} \geq r} p(\tilde{r}).
		\end{equation}
		Here $M$ is the number of estimated bounding boxes, $p(r)$ is the precision as the function of recall $r$, and $p_\text{interp}(r)$ is a smoothed version of the precision curve $p(r)$ \cite{voc2010pascal}.
		The recall value $r_i \in \{r_1,\ldots,r_M\}$ in Equation \ref{eq:ap} is obtained considering the confidence threshold $\tau$ equal to the confidence score of the $i$-th bounding box within the set of estimated bounding boxes when sorted by the confidence score in descending order.
		Throughout this paper we will use \ap{} for varying levels of $\kappa$, denoting it by \ap{} @ IOU $\kappa$.
	
	\subsection{Comparative evaluation of fusion schemes}
	\label{sec:eval:earlyVlate}
		The purpose of this experiment is to compare the performance of early, late and hybrid fusion schemes in terms of their detection performances, communication cost and computation time.
		The detection performance of each scheme is quantified by the \ap{} metric for $\kappa \in \{0.7, 0.8, 0.9\}$.
		The performance evaluation was carried out in both T-junction and roundabout scenarios in this experiment.
		For the late fusion scheme, the NMS algorithm uses an IOU threshold of 0.1, which was experimentally determined to remove multiple detections of a single object.
		For the hybrid fusion scheme, the radius $R$ of the circle limiting the area for low level data sharing was experimentally determined for best trade-off between communication cost and detection performance to be 20m and 12m for the T-junction and roundabout scenarios, respectively.
			
		\begin{table}[]
			\caption{Comparative evaluation of Early Fusion (EF), Hybrid Fusion (HF) and Late Fusion (LF) schemes}
			\label{tab:earlylate}
			\resizebox{\linewidth}{!}{%
			\begin{tabular}{@{}lllllll@{}}
				\toprule
				& & \multicolumn{3}{c}{\textbf{AP\textsubscript{3D}}} & \textbf{Comm. Cost (kbit)} & \textbf{Comp. Time (ms)}  \\  
				\cmidrule(r){3-5} \cmidrule(r){6-6} \cmidrule(r){7-7}	&& $\kappa= 0.7$ & $\kappa= 0.8$ & $\kappa= 0.9$ & \textbf{per sensor}  & \textbf{per frame} \\ \midrule
				\multirow{3}{1.2cm}{\textbf{Tjunction}}  & \textbf{LF} & 0.8181  & 0.6259 & 0.07072 & 0.51 & 298   \\
			                                           	 & \textbf{HF} & 0.8903  & 0.7056 & 0.07277 & 64   & 380   \\ 
				                                         & \textbf{EF} & 0.9870  & 0.9447 & 0.3861  & 516  & 380   \\ \midrule
				                                         				                                         
				\multirow{3}{1.2cm}{\textbf{Roundabout}} & \textbf{LF} & 0.8143  & 0.5986 & 0.01762 & 0.26 & 214   \\
										    			 & \textbf{HF} & 0.8398  & 0.6289 & 0.02013 & 372  & 299   \\
											             & \textbf{EF} & 0.9670  & 0.8816 & 0.04638 & 674  & 299   \\ \bottomrule
			\end{tabular}
		}
		\end{table}

		Table \ref{tab:earlylate} summarises the results of this experiment in terms of the \ap{}, communication cost metrics and computation time for both scenarios.
		These results show that the early fusion scheme outperforms hybrid fusion, which in turn outperforms late fusion.
		More specifically, the early fusion scheme outperforms late fusion scheme up to 20\% in terms of detection performance in the T-junction scenario and 18\% in the roundabout scenario measured by the \ap{} metric with IOU threshold of 0.7.
		The early fusion scheme demonstrates a significantly better detection performance compared to the other schemes when considering higher values of $\kappa$, such as 0.8 and 0.9.
		It can also be seen that for a given value of $\kappa$, the detection performance in the T-junction scenario is consistently superior to the detection performance in the roundabout scenario.
		This arises from the larger voxel sizes that had to be adopted in the latter scenario, for the reasons described in Section \ref{sec:eval:train}, which reduces the spatial resolution of the system and results in less accurate bounding box regression.
		
		The results in Table \ref{tab:earlylate} show that the superiority of the detection performance of the early fusion scheme comes at a higher communication cost.
		This is due to the larger data volume required to transmit raw point clouds in early fusion compared to the transmission of the estimated objects in late fusion from the sensors to the central system.
		Furthermore, the hybrid fusion outperforms late fusion with a significantly lower communication cost than that of early fusion, however underperforms early fusion due to the loss in the omitted points.
		It should be noted that the actual required link capacity from a sensor node to the central fusion system will depend on the processing frame rates.
		For example, using early fusion in the proposed T-junction scenario with a processing frame rate of 10 frames per second, common for lidars, will require a communication link with the capacity of 5.16 Mbps (516 kb per frame times 10 frames per second) from each infrastructure sensor to the central fusion system.
		Such rates can be easily supported by the commercial wired as well as wireless Local Area Network (LAN) technologies that may be needed to implement the proposed system model in Figure \ref{fig:logical}.
		Although we have not considered network delay in this study, our insight is that it could constitute a significant problem to the fusion system by preventing successful detections due to missing frames or by generating false positives due temporal misalignment of incoming frames.
		The likelihood of such miss detections depends on the communication channel properties and should be rigorously investigated in future studies.
				
		The computation time required to compute each frame increases in early fusion because the fused point cloud has more points than the individual point clouds processed separately in the late fusion scheme.
		Note that the computation times are hardware dependent and in this case were obtained using a Nvidia Quadro M4000 GPU.
		
	\subsection{Impact of sensors pose and number on detection performance}
	\label{sec:eval:sconfig}
		This experiment focuses on the evaluation of the impact of the pose (position and orientation) and number of sensors on the object detection performance.
		The performance is evaluated for early and late fusion schemes on both scenarios considering all objects within the detection area, defined in Section \ref{sec:dataset}.
		The evaluation is carried out for all possible sensor sets, where the number of sensors in a set ranges from one to six in the T-junction and one to eight in the roundabout scenario.
		The NMS algorithm and threshold are the same as the previous experiment for consistency of the results.
	
		\begin{table}[]
			\center
			\caption{Detection performance of Early Fusion (EF) and Late Fusion (LF) for various sensor combinations}
			\label{tab:sensornum}
			\resizebox{\linewidth}{!}{%
			\begin{tabular}{@{}clllllll@{}}
				\toprule
				                     &\multicolumn{3}{c}{\textbf{T-junction}}   &\multicolumn{3}{c}{\textbf{Roundabout}} \\ \cmidrule(r){2-4} \cmidrule(r){5-7}
				\textbf{No. Sensors} &\textbf{Sensor Set}  & \textbf{EF \ap}    & \textbf{LF \ap}     &\textbf{Sensor Set}  & \textbf{EF \ap}  &  \textbf{LF \ap} \\ \midrule
				8                    & -                   & -                  & -                   & 0,1,2,3,4,5,6,7     & \textbf{0.9670}  & \textbf{0.8143}  \\ \midrule 
				\multirow{3}{*}{7}   & -                   & -                  & -                   & 0,1,2,3,5,6,7       & \textbf{0.9385}  & \textbf{0.7904}  \\ 
				                     & -                   & -                  & -                   & 1,2,3,4,5,6,7       & 0.9340           & 0.7868           \\
				                     & -                   & -                  & -                   & 0,1,3,4,5,6,7       & 0.9307           & 0.7834           \\ \midrule
				\multirow{3}{*}{6}   & 0,1,2,3,4,5         & \textbf{0.9870}    & \textbf{0.8181}     & 1,2,3,5,6,7         & \textbf{0.9050}  & 0.7627           \\ 
				                     & -                   & -                  & -                   & 1,2,3,4,5,7         & 0.9017           & 0.7627           \\
				                     & -                   & -                  & -                   & 0,1,2,3,5,7         & 0.8973           & \textbf{0.7664}  \\ \midrule
				\multirow{3}{*}{5}   & 0,1,3,4,5           & \textbf{0.9441}    & 0.7611              & 1,2,3,5,7           & \textbf{0.8678}  & \textbf{0.7385}  \\ 
				                     & 0,1,2,3,5           & 0.9348             & 0.6672              & 1,3,5,6,7           & 0.8610           & 0.7314           \\
				                     & 1,2,3,4,5           & 0.9327             & \textbf{0.7914}     & 1,3,4,5,7           & 0.8576           & 0.7313           \\ \midrule
				\multirow{3}{*}{4}   & 1,3,4,5             & \textbf{0.8653}    & \textbf{0.7336}     & 1,3,5,7             & \textbf{0.8231}  & \textbf{0.7069}  \\ 
				                     & 0,1,2,5             & 0.8596             & 0.4765              & 1,2,3,7             & 0.7356           & 0.6557           \\
				                     & 0,1,3,4             & 0.8394             & 0.6827              & 1,3,4,7             & 0.7263           & 0.6482           \\ \midrule
				\multirow{3}{*}{3}   & 0,2,5               & \textbf{0.7123}    & 0.4039              & 1,3,7               & \textbf{0.6892}  & \textbf{0.6230}  \\ 
				                     & 2,3,5               & 0.6938             & 0.5834              & 3,5,7               & 0.6713           & 0.5831           \\
				                     & 3,4,5               & 0.6837             & \textbf{0.6656}     & 1,3,5               & 0.6382           & 0.5861           \\ \midrule
				\multirow{3}{*}{2}   & 3,4                 & \textbf{0.5365}    & \textbf{0.5361}     & 3,7                 & \textbf{0.5016}  & 0.4594           \\ 
				                     & 2,3                 & 0.4591             & 0.4530              & 1,3                 & 0.4995           & \textbf{0.4998}  \\
				                     & 2,5                 & 0.4453             & 0.3309              & 5,7                 & 0.4664           & 0.4638           \\ \midrule	
				\multirow{3}{*}{1}   & 4                   & \textbf{0.2862}    & \textbf{0.2862}     & 3                   & \textbf{0.2811}  & \textbf{0.2811}  \\ 
				                     & 3                   & 0.2512             & 0.2512              & 5                   & 0.2596           & 0.2596           \\
				                     & 2                   & 0.2013             & 0.2013              & 1                   & 0.2151           & 0.2151           \\ \bottomrule 
			\end{tabular}
			}
		\end{table}	
	
		\begin{figure*}
			\centering
			\subfloat[T-junction \label{fig:pr-t}]{\includegraphics[width=0.48\textwidth]{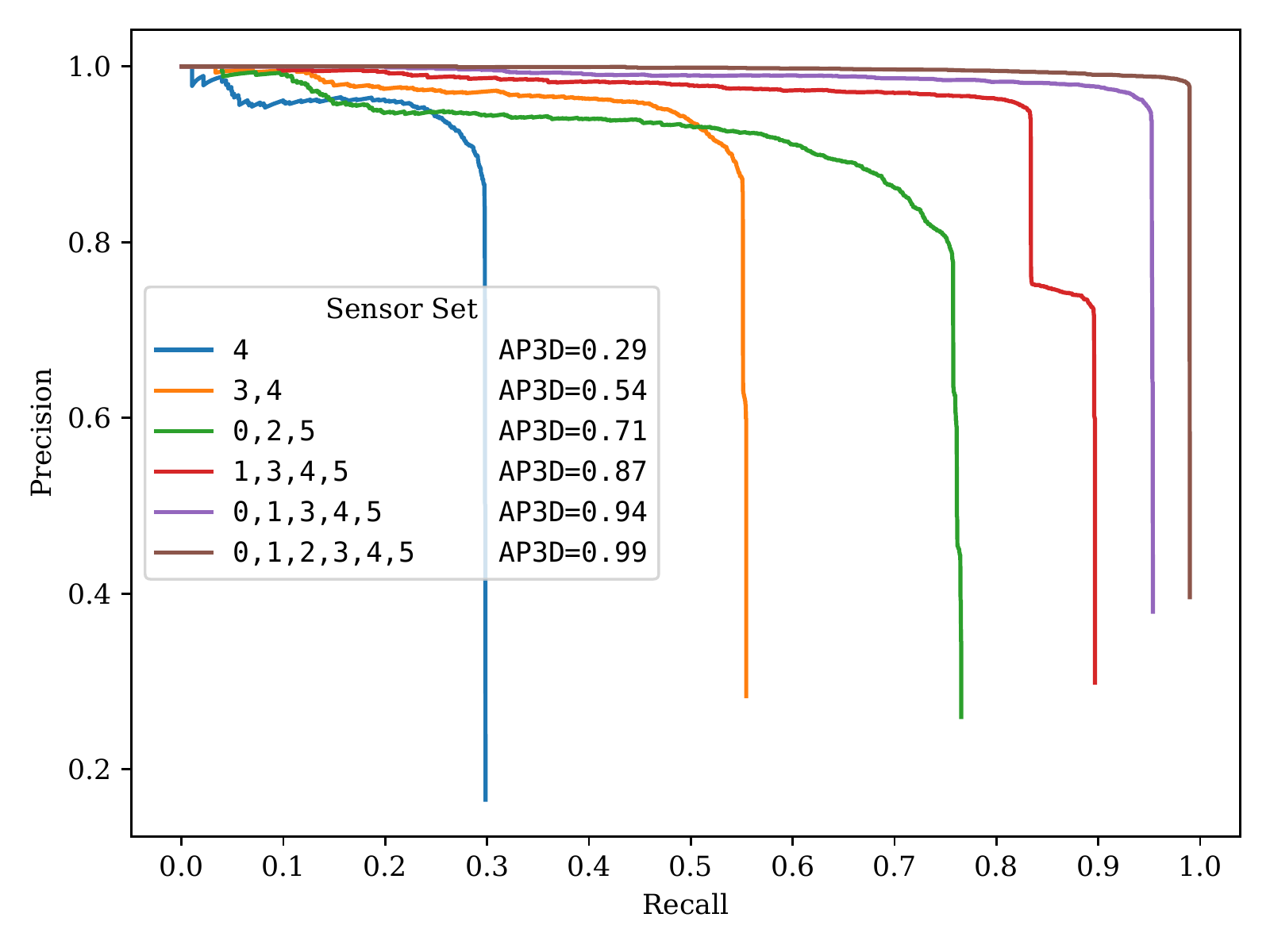}}\hfill
			\subfloat[Roundabout \label{fig:pr-r}]{\includegraphics[width=0.48\textwidth]{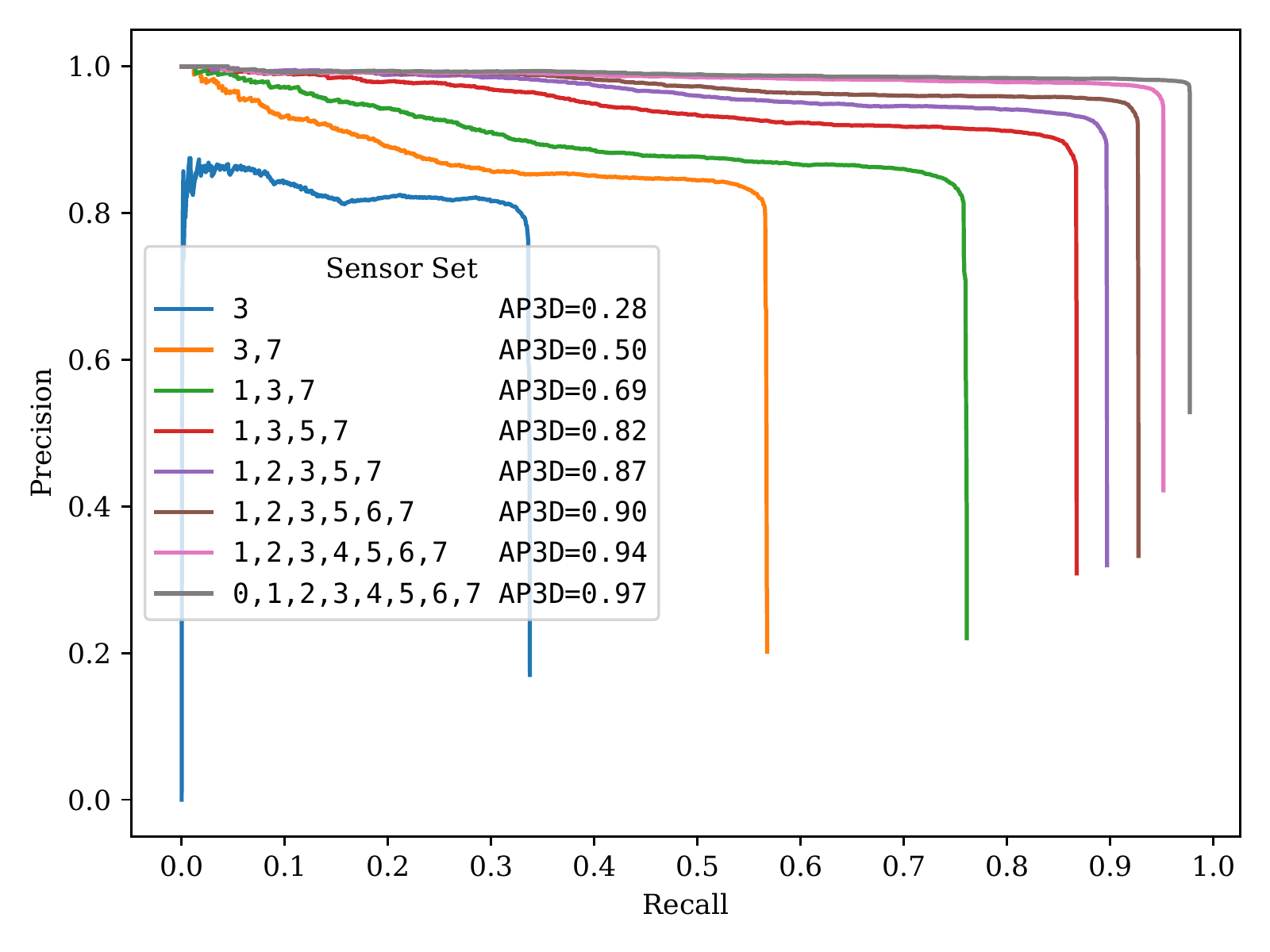}}
			
			\caption{Precision-Recall curves of early fusion with different number of sensors for \protect\subref{fig:pr-t} T-junction and \protect\subref{fig:pr-r} roundabout scenarios. The curves are produced for the sensor sets highlighted in bold in Table \ref{tab:sensornum}. \ap{} values are calculated for $\kappa=0.7$.} 
			\label{fig:pr}
		\end{figure*}
		
		Table \ref{tab:sensornum} reports the top-3 performing sensor sets for each number of sensors in both scenarios in terms of \ap{} metric ($\kappa=0.7$).
		The results show that the detection performance increases as the number of engaged sensors is increased.
		In particular, there is a steep performance increase of more than 50\% when using two sensors instead of one in both scenarios.
		However, the performance gain saturates as the number of sensors increases.
		Also, it can be seen that as the detection area increases, more sensors are needed to maintain the detection performance.
		For example, in the roundabout scenario, eight sensors need to be engaged to achieve a performance level equal to that of six engaged sensors in the T-junction scenario, which has smaller detection area.
		Furthermore, the early fusion scheme consistently outperforms late fusion with respect to the detection performance.
		Their disparity becomes more significant as the number of sensors grow since the early fusion scheme can exploit more information at detection time compared to late fusion.
				
		Figure \ref{fig:pr} presents the PR curves of the best sensor sets (rows with a bold font in Table \ref{tab:sensornum}) for both scenarios using the early fusion scheme.
		The curves show that the maximum recall of detected objects increases significantly for both scenarios when the number of engaged sensors increase.
		Specifically, a single sensor can detect only 30\% of all the vehicles in the T-junction scenario and slightly more than 30\% of all the vehicles in the roundabout scenario.
		However, when all sensors (six for the T-junction and eight for the roundabout) are engaged, both scenarios show similar performance and detect more than 95\% of the ground-truth objects with precision above 95\%.
		
		As one could anticipate, the results show that the number and pose of sensors have direct impact on the performance of the system.
		The results also demonstrate the impact of spatial diversity in terms of the improved quality of the input information to the object detection model.
		For example, the sensor set (0,2,5) achieve the best performance for three sensors in the T-junction scenario.
		Adding sensor 1 to the aforementioned set does not increase the field-of-view of the system, as observed in Figure \ref{fig:config-t}, however the results in Table \ref{tab:sensornum} show that this addition has a notable impact on detection performance (20\% increase in \ap{}).
		The impact of spatial diversity on detection performance is further explored in the next experiment.
	
	\subsection{Spatial diversity gain of cooperative perception}
	\label{sec:eval:overlap}
		The enhanced detection performance in cooperative perception seen in the previous experiments can be associated with two factors: 1) the increased field-of-view; 2) the spatial diversity gain, which manifests itself in point clouds with higher point density in areas that are covered by multiple sensors.
		This experiment intends to shed light into the latter factor.
		
		In this experiment, we focus on objects within a defined Region of Interest (ROI), where all objects are within the field-of-view of two specific sensors.
		For example, the ROI for the sensor set (2,4) in the T-junction scenario is limited to road to the left of the junction (filled with green and brown points in Figure \ref{fig:pcl-t}), and the ROI for the sensor set (3,5) for the roundabout is limited to the upper right quadrant of the roundabout (filled with yellow and light purple points in Figure \ref{fig:pcl-r}).
		For each of these ROIs, the detection performance of the early fusion scheme using the specified sensor sets is compared to that of a single sensor covering the same ROI.
		The detection performance is quantified by the \ap{} metric restricted to objects within the specified ROI.
		For each scenario we considered the two sensors sets with highest field-of-view overlap: (1,5) and (2,4) for the T-junction scenario and (1,7),(3,5) for the roundabout scenario.
				
		\begin{table}[]
			\center
			\caption{Impact of spatial diversity on detection performance}
			\label{tab:overlap}
			\begin{tabular}{@{}clllllll@{}}
				\toprule
				\multicolumn{3}{c}{\textbf{T-junction}}                       & \multicolumn{3}{c}{\textbf{Roundabout}} \\ \cmidrule(r){1-3} \cmidrule(r){4-6}
				\textbf{ROI}          & \textbf{Sensors Set} & \textbf{\ap}   & \textbf{ROI}         & \textbf{Sensors Set} & \textbf{\ap}           \\ \midrule
				\multirow{3}{*}{1,5}  & 1                    & 0.4717         & \multirow{3}{*}{1,7} & 1                    & 0.1474   \\ 
				                      & 5                    & 0.3222         &                      & 7                    & 0.8874   \\
				                      & 1,5                  & 0.8722         &                      & 1,7                  & 0.8925   \\ \midrule
				\multirow{3}{*}{2,4}  & 2                    & 0.5621         & \multirow{3}{*}{3,5} & 3                    & 0.3944   \\ 
									  & 4                    & 0.7942         &                      & 5                    & 0.8819   \\
									  & 2,4                  & 0.9560         &                      & 3,5                  & 0.8996   \\ \midrule
			\end{tabular}
		\end{table}
	
\begin{figure*}
			\centering
			\subfloat[From Sensor 0]{\includegraphics[width=0.24\textwidth]{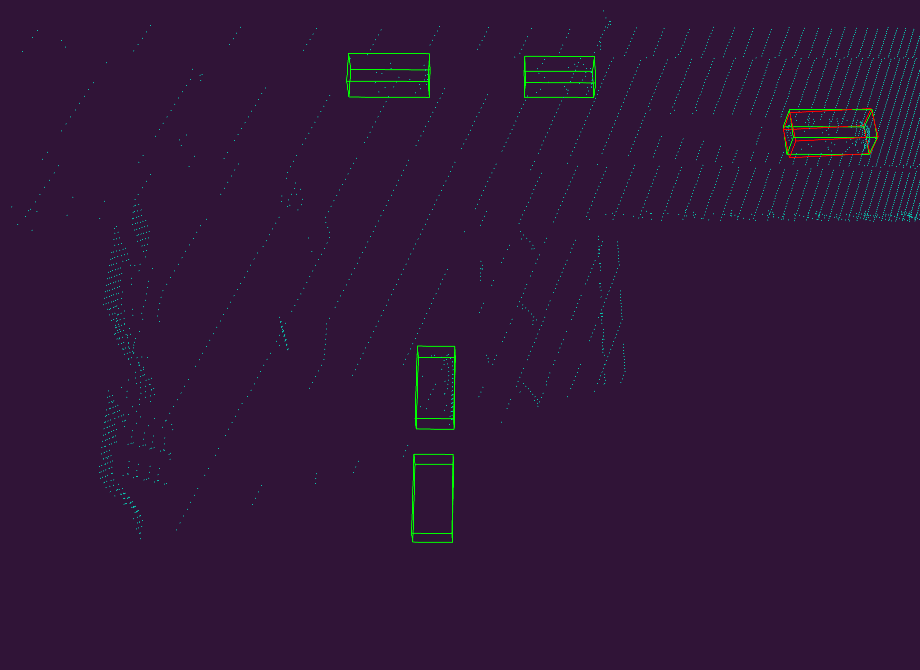}} \hfill
			\subfloat[From Sensor 1 \label{fig:demo:1}]{\includegraphics[width=0.24\textwidth]{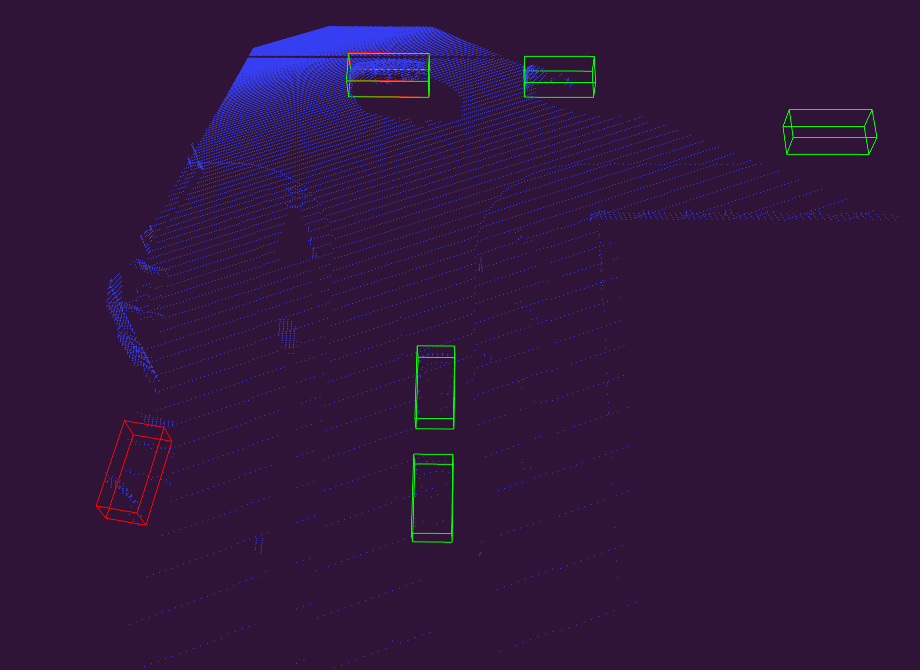}} \hfill			
			\subfloat[From Sensor 5 \label{fig:demo:5}]{\includegraphics[width=0.24\textwidth]{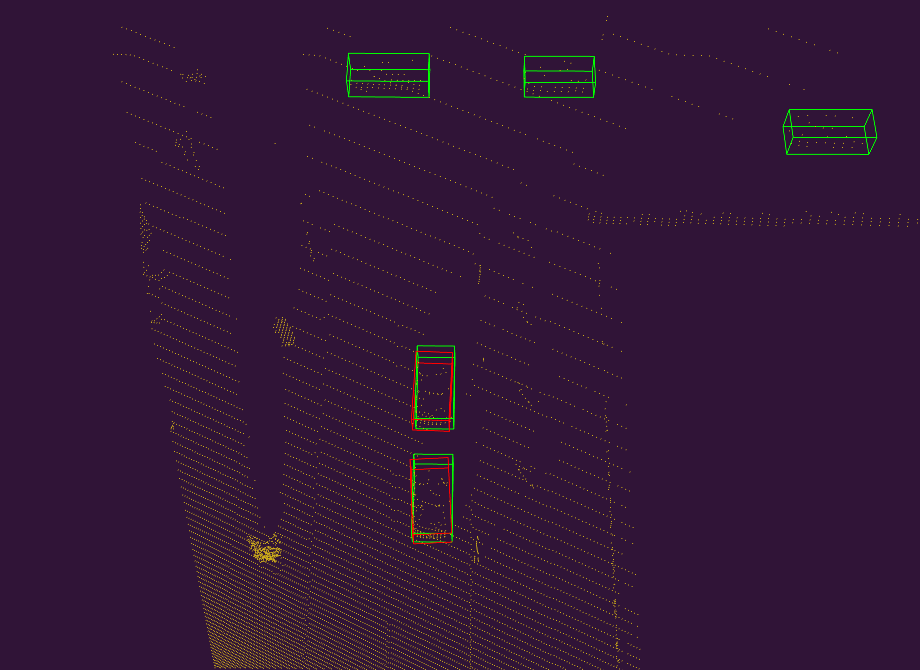}} \hfill
			\subfloat[Early Fusion from sensor set (0,1,5) \label{fig:demo:015}]{\includegraphics[width=0.24\textwidth]{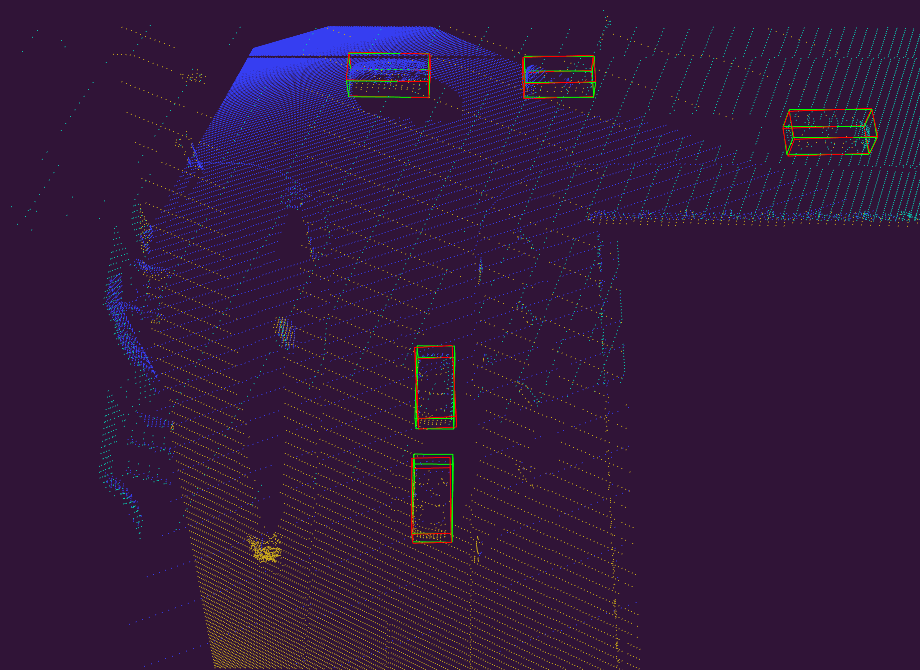}}			
			\caption{Illustration of the impact of spatial diversity on the performance of the early fusion scheme for various sensor configurations (green boxes represent the ground-truth objects and red boxes represent estimated objects).} 
			\label{fig:demo}
		\end{figure*}

		The impact of spatial diversity on the detection performance of early fusion scheme is visualised in Figure \ref{fig:demo}.
		As it can be seen in the snapshots in Fig \ref{fig:demo:5}, when a single sensor is engaged most objects fail to be detected or are detected with poor accuracy (\textit{i.e.} incorrect size or yaw angle).
		However, upon increasing the point density by combining multiple point clouds with early fusion, it is possible reduce the number of false negatives and increase the quality of estimated bounding boxes, as illustrated in Figure \ref{fig:demo:015}.
	
		The results of this experiment, summarised in Table \ref{tab:overlap}, show that the early fusion scheme using only two sensors outperforms the best single sensor by 20\% and 85\% in the T-junction scenario.
		However, the detection performance gain when using two sensors in the roundabout is marginal.
		This is due to the fact that the fields-of-view of the specific sensor set used has minimal overlap in this particular roundabout scenario.		
		The results presented indicate that early fusion can: a) reduce the number of false negatives caused by occlusion and low point density; b) improve the quality of estimated boxes when the sensors have significant overlapping coverage.
		
	\subsection{Impact of point density on estimated bounding boxes accuracy}
	\label{sec:eval:cov}	
		One can intuitively stipulate that a denser point cloud will provide additional information about the objects in the scene, as has been the case for Airborne lidar scanners and object classification in context of remote sensing \cite{tomljenovic2014influence}.
		This experiment analyses how the number of sensors affects the density of points in the point cloud and, in turn, the accuracy of the boxes estimated by the object detection model in a driving context.
		We define the point density of an object as the number of points within the boundaries of its ground-truth bounding box.
		The point density of an object is a discrete random variable that is a random function of the number and pose of sensors that observe the object.
		
\begin{figure*}
			\centering
			\subfloat[CDF of objects point density for a varying number of sensors. An object's point density is the number of points within its ground-truth box. The curve slope represent the number of objects with a specific point density.
			\label{fig:exp2:ptshist}]{\includegraphics[width=0.48\textwidth]{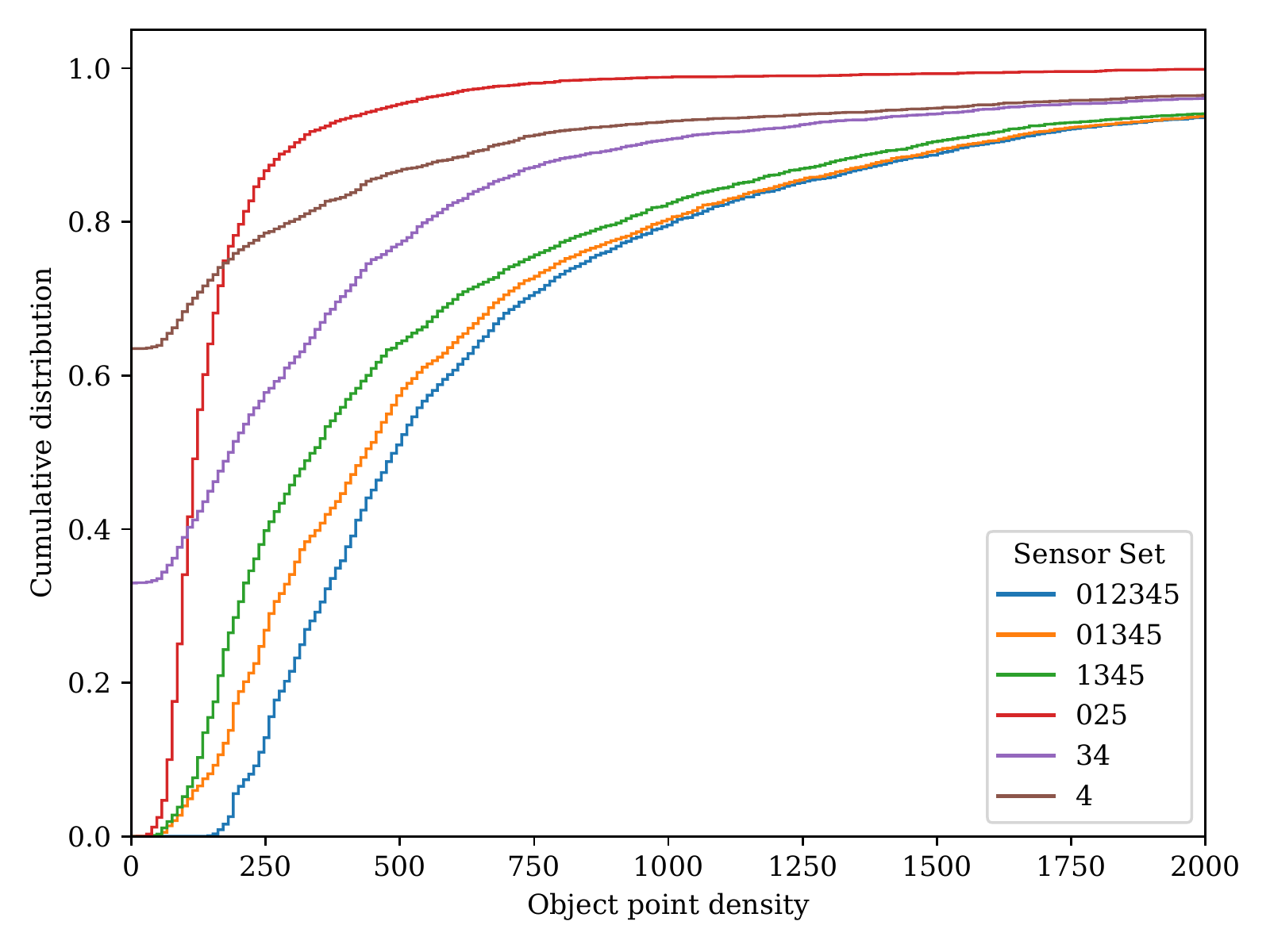}}\hfill
			\subfloat[IOU between ground-truth and detected boxes as the object point density varies. Each point represents the average IOU for a bin of objects. The number of bins used in the interval was 200. \label{fig:exp2:iou}]{\includegraphics[width=0.48\textwidth]{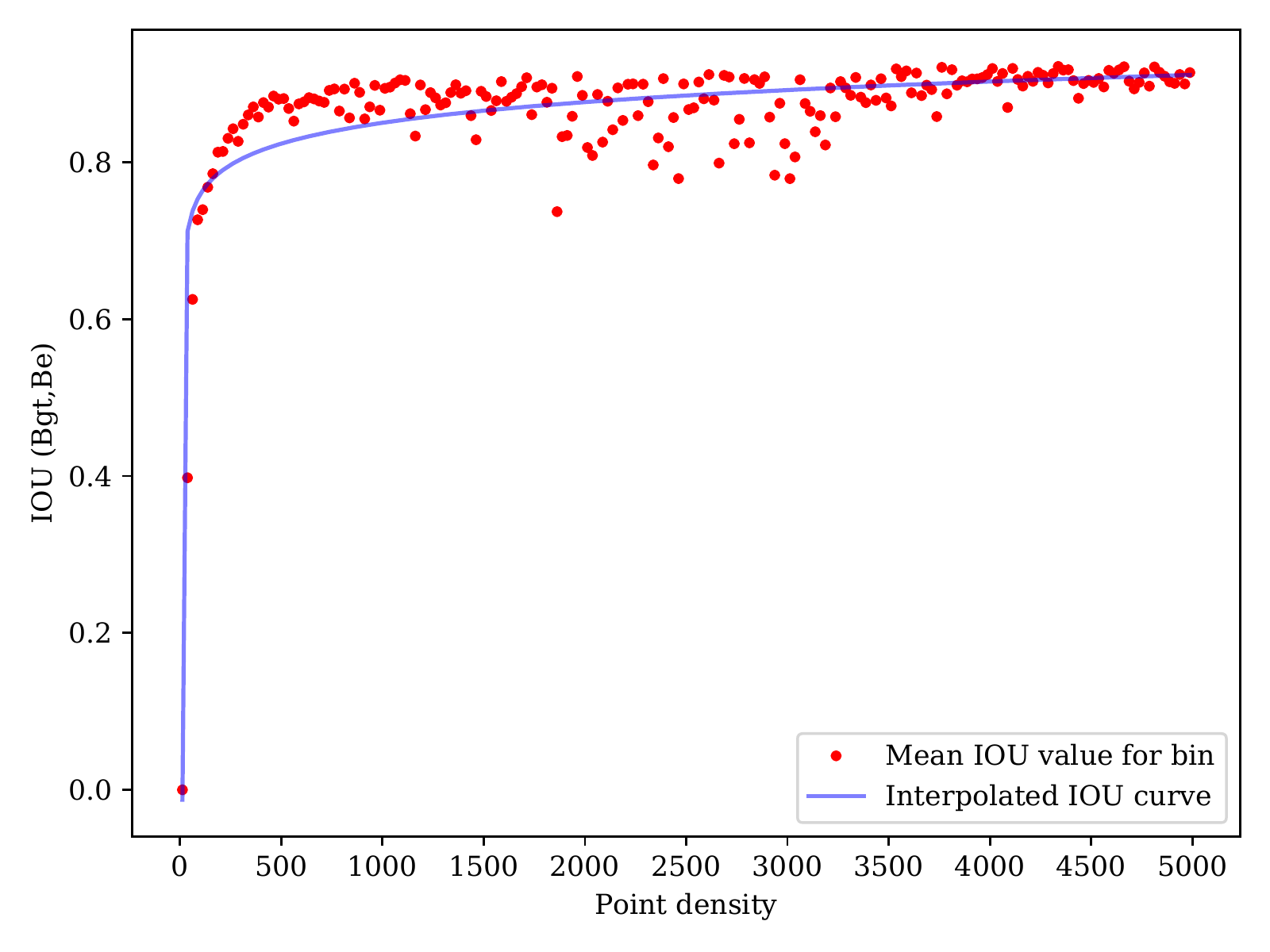}}
			\caption{Analysis of the point density and the IOU metric over estimated boxes in the T-junction test dataset.}
			\label{fig:exp2}
		\end{figure*}

		Figure \ref{fig:exp2:ptshist} shows the Cumulative Distribution Function (CDF) of objects' point density for the best sensor sets in the T-junction scenario from Table \ref{tab:sensornum} (highlighted in bold font).
		Given a point $(d, F(d))$ where $F$ represents one of the CDF curves, the vertical coordinate $F(d)$ represents the ratio of objects whose point density is smaller or equal to the horizontal coordinate $d$.
		The intersections with the vertical axis shows that using Sensors 4 and the sensor set (3,4) alone results in more than 60\% and 30\% of the objects having zero point density, respectively.
		Similarly, one can compute the ratio of objects whose point density is within an interval $[d_1,d_2]$ by computing $F(d_2)-F(d_1)$.
		Thus, using the sensor set (0,2,5) guarantees that all ground-truth objects have non-zero point density but results in 90\% of the objects having point density below 300 points.
		When the number of engaged sensors is increased, the number of objects with point density in the range of $[250,1000]$ points increases significantly, but saturates for point densities above 1750 points.

		Next, we investigate how object point density relates to the accuracy of the estimated bounding box, measured by the IOU metric.
		For this purpose, we split the objects into 200 uniform-sized bins according to point density.
		The IOU value for a bin is computed by averaging the individual IOU values among all objects in the bin.
		Figure \ref{fig:exp2:iou} shows the scatter plot of the IOU value per point density bin and a log curve interpolation.
		The accuracy of objects with point density below 70 points is poor, but increases significantly when the point density surpasses 100 points.
		The outliers observed in the range of $[1800,3400]$ are caused by objects that are close to a sensor, thus have a high number of points concentrated on a small surface but few points elsewhere, resulting in a poorly estimated bounding box.
		In conclusion, the point density of an object can provide a useful prediction of the accuracy of the estimated bounding box.
		Thus, given the accuracy requirement for the estimated bounding boxes, it is possible to find the minimum required point density and the number of sensors required for a specific scenario.
		
	\subsection{Comparison with existing benchmarks}
	\label{sec:eval:comp}
	
		\begin{table}[]
			\caption{Comparison with existing benchmarks}
			\label{tab:benchmarks}
			\resizebox{\linewidth}{!}{%
				\begin{tabular}{@{}lll@{}}
					\toprule
					 & \multicolumn{2}{c}{\textbf{AP\textsubscript{3D}}}  \\ \cmidrule(r){2-3}
					 &  \textbf{Single sensor} & \textbf{Two sensors} 						  \\ \midrule  
					Voxelnet \cite{voxelnet}          & 0.8197(E), 0.6546(M), 0.6285(H) & -   \\ 
					Cooper   \cite{chen_cooper_2019}  & 0.1960 & 0.7237  \\
					F-Cooper \cite{fcooper}           & 0.1960 & 0.7237  \\
					Ours (early fusion)               & 0.4717 & 0.8722  \\ \bottomrule
				\end{tabular}
			}
		\end{table}
	
		The direct comparison of our fusion schemes with other approaches \cite{voxelnet,fcooper} may not be meaningful due to its unique sensing strategy.
		However, we opt to compare (a) the \ap{} results obtained using a single sensor to the reported results produced by Voxelnet \cite{voxelnet}; (b) the \ap{} results using two sensors in the T-junction scenario from Section \ref{sec:eval:overlap} to the reported results produced by Cooper \cite{chen_cooper_2019} and F-Cooper \cite{fcooper} in a road intersection scenario in Table \ref{tab:benchmarks}.
		
		Firstly, we compare results produced by Voxelnet \cite{voxelnet} using the KITTI dataset \cite{kitti}. 
		The results reported for \ap{} in three complexity categories, easy, moderate and hard, are 81.97\%, 65.46\% and 62.85\%, respectively.
		Our results from Section \ref{sec:eval:sconfig} using a single sensor achieve much lower \ap{}, around 28\% for both scenarios.
		This significant performance gap emerges due to our evaluation considering all the ground-truth objects within the detection area, while Voxelnet and other studies using the KITTI benchmark consider only the objects within the sensor's field-of-view.
		The performance gap highlights the complexity of detecting objects in both scenarios using a single sensor, since its field-of-view cannot cover all the detection area and is susceptible to occlusion caused by buildings and other objects.
		As discussed in Section \ref{sec:eval:sconfig}, increasing the number of sensors used is highly beneficial to the detection performance in the proposed system.
		
		Secondly, we compare our early fusion scheme results with the ones produced in F-Cooper \cite{fcooper}.
		For a fair comparison, we consider our system using two engaged sensors in the T-junction scenario to the ``road intersections'' scenario reported in \cite{fcooper}.
		In \cite{fcooper}, the authors report results in two categories, ``near and far'', according to the distance from the object to the sensor. 
		The ``near'' category shows marginal improvement, hence we focus the comparison on the ``far'' category.
		The \ap{} results in \cite{fcooper} for a single sensor and fusion of two sensors are 19.60\% and 72.37\%, respectively.
		Our results in Section \ref{sec:eval:overlap} under similar scenario for a single sensor and early fusion of two sensors are 47.17\% and 87.22\%, respectively.
		Although the direct comparison of these values is not meaningful given the dataset differences and sensing strategy, it is possible to see that both approaches show a comparable performance gain when considering more than a single sensor.
				
\section{Conclusion}
\label{sec:conclusion}
	This paper proposed a cooperative perception system for 3D object detection using two fusion schemes: early and late fusion.
	The proposed system model contains $n$ infrastructure sensors that share data with a central fusion system, where information is fused and the resulting detections (3D bounding boxes) are disseminated to all the vehicles in the vicinity.
	A novel cooperative dataset containing depth maps from multiple infrastructure sensors in a T-junction and a roundabout scenario was used for the evaluation of the proposed system.
	The evaluation indicated that increasing the number of sensors in the proposed system is highly beneficial in complex scenarios, which allowed to overcome occlusion and restricted field-of-view.
	Furthermore, the proposed system was able to increase the density of the fused point cloud by exploiting spatially diverse observations with overlapping fields-of-view, which reduced false negative detections and allowed more accurate estimation of bounding boxes.		
	Finally, the results suggested that the system can be realised with current communications technologies and can reduce the costs of individual vehicles through shared infrastructure resources.

	Future research opportunities include the investigation of cooperative perception using vehicles' sensor data, where localisation estimation as well as bandwidth requirements can be more challenging.
	We also envisage more efficient data fusion schemes, where the transferred data volume can be reduced while maintaining the 3D object detection performance levels.